%% file: main.tex
\documentclass[letterpaper]{article} 
\usepackage[preprint]{aaai2027}  
\usepackage[hyphens]{url}  
\usepackage{graphicx} 
\usepackage{natbib}  
\usepackage{caption} 
\usepackage{booktabs}
\usepackage{algorithm}
\usepackage{amsmath}
\usepackage{amssymb}
\usepackage{amsthm}
\usepackage{algorithmic}
\usepackage{tikz}
\usetikzlibrary{arrows.meta,positioning}

\usepackage{newfloat}
\usepackage{listings}
\DeclareCaptionStyle{ruled}{labelfont=normalfont,labelsep=colon,strut=off} 
\floatstyle{ruled}
\newfloat{listing}{tb}{lst}{}
\floatname{listing}{Listing}

\usepackage{booktabs}

\title{Question Begets Question: Self-Evolving Curriculum\\ for Reinforcement Fine-Tuning on Competition Mathematics}

\author{
    Longtian Bao\textsuperscript{\rm 1}\equalcontrib,
    Jianyou Wang\textsuperscript{\rm 2}\equalcontrib,
    Yang Zhang\textsuperscript{\rm 2}\equalcontrib,
    Youze Zheng\textsuperscript{\rm 2}\equalcontrib,
    Ramamohan Paturi\textsuperscript{\rm 2}
}
\affiliations{
    \textsuperscript{\rm 1}University of Chicago\\
    \textsuperscript{\rm 2}University of California, San Diego\\
    lobao@uchicago.edu, \{jiw101, yaz124, yoz018, rpaturi\}@ucsd.edu
}

\begin{document}

\maketitle

\begin{abstract}
\input{sections/abstract}
\end{abstract}

\begin{figure*}[t]
    \centering
    \includegraphics[width=0.99\textwidth]{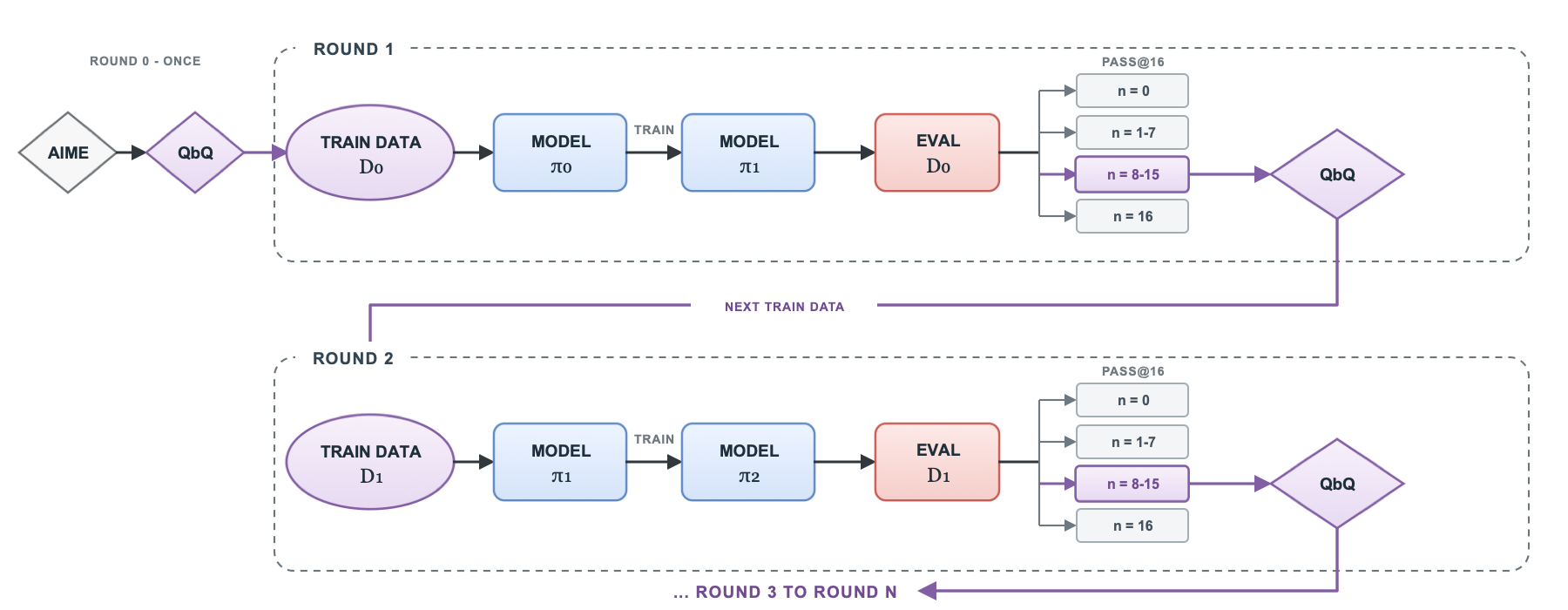}
    \caption{Overview of QbQ. The initial AIME training set is created once. In each subsequent round, the model is trained and evaluated, problems are grouped by pass@16, and only those with ($8 \leq n \leq 15$) are used to generate training data for the next round.}
    \label{fig:qbq_overview}
\end{figure*}
\input{sections/introduction}

\input{sections/related_work}
\input{sections/method}

\input{sections/experiment}
\input{sections/conclusion}

\section*{Acknowledgments}
We gratefully acknowledge Professor Chenhao Tan and the University of Chicago Data Science Institute for providing access to the DSI computing cluster. The computational resources made available through the cluster were instrumental in supporting the experiments and analyses conducted in this work.

\bibliography{aaai2027}

\clearpage
\appendix
\input{sections/appendix}


\end{document}

%% file: sections/abstract.tex
Teaching a language model a skill it has not mastered is obstructed by three recurring difficulties: training data is scarce, ground-truth reasoning traces are usually unavailable, and models often exhibit an apparent ceiling beyond which additional data yields no further improvement. We study these difficulties in a controlled setting, fine-tuning Qwen2.5-Math-7B on competition mathematics
(AIME), a task on which it initially solves only 5.6\% of problems (pass@1). To address data scarcity, we introduce Question-begets-Question (QbQ), a scalable procedure in which a teacher transforms existing problems into diverse variants that probe the same underlying skills; to model the absence of oracle reasoning,
we train exclusively via reinforcement learning on problem statements and final answers, never on teacher reasoning traces. Static training on such data, however, plateaus well short of the task: real-plus-synthetic augmentation and non-curriculum QbQ generated synthetic data training cap pass@1 at 12.5\% and 14.5\% respectively, despite large increases in data. Our central finding is that this ceiling is not intrinsic to the model. We propose a self-evolving curriculum that, each round, evaluates the current checkpoint, seeds QbQ from the problems it can mostly get
right, and trains on the resulting variants; under an identical data budget, this breaks the ceiling and lifts pass@1 to 16.5\% with no sign of saturation after 20 rounds. Counterintuitively, we find that models improve when trained on variants of problems they can mostly get right, and that models trained this way go on to solve harder problems never seen during training.

%% file: sections/introduction.tex

\section{Introduction}
\label{sec:introduction}
Teaching language models to acquire new skills is a fundamental problem in
modern AI research. Although frontier large language models now match or exceed
human experts in many domains, they continue to lag behind in others, such as
constructing lengthy mathematical proofs \citep{petrov2025proofbluffevaluatingllms}
or interpreting and litigating complex legal documents
\citep{guha2023legalbenchcollaborativelybuiltbenchmark,dahl2024largelegalfictionsprofiling}.
On tasks that challenge even human experts, including drug discovery
\citep{guo2026largermodelsreallywin}, clinical-trial outcome prediction
\citep{wang2026ctopenopenaccessuncontaminated}, earthquake forecasting
\citep{stockman2026earthquakenppbenchmarkearthquakeforecasting}, and political
forecasting \citep{karger2025forecastbenchdynamicbenchmarkai}, these models fare
little better.

We attribute this gap to three recurring obstacles. First, for difficult tasks,
training data is often scarce: M7+ earthquake catalogs, for instance, contain
fewer than 2{,}000 usable records \citep{storchak2013iscgem}. Second, even when
data is abundant, correct reasoning traces are rarely available. Clinical-trial
outcome prediction offers on the order of 500{,}000 labeled examples, yet none
come with ground-truth explanations of why a trial succeeded or failed, and
expert rationales in this setting are themselves often flawed and largely post
hoc \citep{wang2026ctopenopenaccessuncontaminated}. Third, and most important, a
model may face an apparent learning bottleneck, an ostensible ceiling on what it
can master regardless of the amount of training data available
\citep{yue2025doesreinforcementlearningreally}.

We address each obstacle in turn. To overcome data scarcity, we present a
method, Question-begets-Question (QbQ), that generates an abundance of synthetic
data in a scalable manner while improving its diversity. To mirror settings in
which oracle ground-truth reasoning traces are unavailable, we train our models
on this synthetic data purely through reinforcement learning, never exposing
them to reasoning traces. Finally, to test whether the ostensible ceiling can be
broken under a fixed data budget, we propose an iterative, self-evolving
curriculum that dynamically tailors training data to the evolving needs of the
model, so that it may learn more effectively and surpass the ceiling.

While we lack the resources to study frontier models on the hardest open
problems, we instead study a smaller model on a task that is genuinely hard for
it. Concretely, we fine-tune Qwen2.5-Math-7B (released in 2024) on high-school
competition mathematics (AIME). This is an appropriate testbed: the base model
solves only about 5.6\% of problems (pass@1) on the AIME 2025 and 2026
benchmarks, placing the task well beyond its current reach.

The natural remedies are only partly effective. Using the entire set of 1{,}005
problems with annotated solutions available prior to AIME 2024, supervised
fine-tuning and GRPO raise pass@1 to 9.5\% and 11.5\%, respectively. Standard
synthetic-data augmentation helps only modestly. When we use a stronger teacher,
GPT-5-mini, to generate an additional 5{,}000 problems from the original
1{,}005, training on the combined 6{,}000 examples raises pass@1 to just 12.05\%
despite the sixfold increase in data. We
partly attribute this limited benefit to a lack of diversity among the generated
problems.

QbQ generates markedly more diverse synthetic data than standard augmentation,
and training on 6{,}000 QbQ problems alone, with no real data, already improves
performance to 14.5\%. Yet the ostensible ceiling reappears Figure 2: over the final
30\% of GRPO training, performance plateaus and does not improve as more data is
consumed. Our central result is that this ceiling is not intrinsic to the model.
When the same QbQ data is generated, organized, and trained upon within our
self-evolving curriculum, and under the same data budget, the model breaks
through to 16.5\% pass@1, with no sign of saturation even after 20 rounds. The entire workflow is shown in Figure \ref{fig:qbq_overview}.

Our self-evolving curriculum runs for 20 rounds, each using GPT-5-mini to
synthesize problems tailored to the current checkpoint. Each round has three
stages. (1) \emph{Self-evaluation}: the current checkpoint is evaluated on the
previous round's synthetic problems and sorts them into four bands, problems it
has mastered ($\text{pass@}16 = 16$), problems it can mostly get right
($8 \leq \text{pass@}16 \leq 15$), problems it sometimes gets right
($1 \leq \text{pass@}16 \leq 7$), and problems it never gets right, which are
too hard for it. (2) \emph{Generation}: taking the problems it can mostly get
right as seeds, the teacher synthesizes variants that probe a slightly different
aspect of the same problem, reinforcing rather than escalating its
understanding. (3) \emph{Training}: we apply GRPO to the newly generated
problems using only their questions and final answers, with no step-by-step
solutions. Because the checkpoints never see or train on the teacher's reasoning
traces, this both models the no-oracle-reasoning setting and mitigates concerns
about distillation.

The approach has two key strengths. Because each round's problems are seeded
from the previous round's rather than from a fixed set, the diversity collapse
that hampered static augmentation does not arise. And because problems are
continually re-tailored to the current checkpoint, the model always trains on
the most productive material, variants of problems it can mostly get right,
which improves the effectiveness of GRPO.

This last point underlies our central and counterintuitive finding. Whereas
prior work argues that models improve by learning from their mistakes or
hardest failures \citep{liang2025swsselfawareweaknessdrivenproblem,chen2026nudgingboundariesllmreasoning,chen2026cogdriftexplorationadaptivelyreformulated}, we observe the opposite: a model
improves most when trained on variants of problems it can mostly get right.
Seeding from these near-successes at each round consistently outperforms seeding
from the hardest problems. Perhaps most striking, a model trained under this
curriculum goes on to solve harder problems it was never trained on. All
experiments are conducted under matched compute and data budgets unless
otherwise specified.

\paragraph{Contributions.} Our contributions are as follows:
\begin{itemize}
    \item We propose \emph{Question-begets-Question} (QbQ), a scalable method
    for generating diverse synthetic training problems, and train on it purely
    via reinforcement learning without any teacher reasoning traces, modeling
    settings where ground-truth reasoning is unavailable.
    \item We document an ostensible performance ceiling under static training:
    both real-plus-synthetic augmentation and non-curriculum QbQ training plateau as
    data increases, capping pass@1 at 12.5\% and 14.5\% respectively.
    \item We introduce a self-evolving iterative curriculum that breaks this
    ceiling under an identical data budget, lifting Qwen2.5-Math-7B from 5.6\%
    to 16.5\% pass@1 with no sign of saturation after 20 rounds.
    \item We present a counterintuitive training principle: models improve most
    from variants of problems they can \emph{mostly} get right, not from their
    hardest failures, and models trained this way generalize to harder problems
    never seen in training.
\end{itemize}

%% file: sections/related_work.tex
\section{Related Work}
\label{sec:related}
\paragraph{Competition Mathematics Benchmarks:}
Mathematical benchmarks for language models span a wide range of difficulty.
GSM8K and MATH-500 sit at the easier end
\citep{cobbe2021trainingverifierssolvemath,hendrycks2021measuringmathematicalproblemsolving,lightman2023letsverifystepstep},
and current reasoning models perform well on both. At the hard end, Putnam,
USAMO, and IMO problems typically demand written proofs, which are difficult to
grade automatically
\citep{tsoukalas2024putnambenchevaluatingneuraltheoremprovers,aops2024usamo,imofoundation2024imo}.
Among competition benchmarks whose final answers are computation-centric, AIME
is one of the hardest \citep{aops2024aime}. FrontierMath reaches further still,
into research mathematics \citep{glazer2025frontiermathbenchmarkevaluatingadvanced},
but its problems were constructed to test AI systems rather than people. In this
work, we focus on problems that have a life beyond AI evaluation. AIME problems
are written for human competitors, every answer is an integer that a program can
check, and four decades of contests supply more than a thousand problems in a
single uniform format (\S\ref{sec:setup}). We therefore train and evaluate on
AIME.

\paragraph{Curriculum Learning.}
Curriculum learning organizes training examples according to the learner's current ability. In reinforcement fine tuning, this often means selecting problems that are neither already mastered nor completely beyond the model's reach, since these two cases provide limited learning signal \citep{shi2026efficientreinforcementfinetuningadaptive,mahrooghi2026goldilocksrltuningtask}. Recent work has explored adaptive task selection, problem reformulation, and training near the model's current capability boundary \citep{chen2025selfevolvingcurriculumllmreasoning,chen2026cogdriftexplorationadaptivelyreformulated,sundaram2026teachingmodelsteachthemselves,lee2026zoneproximalpolicyoptimization}. Our method follows the same general idea but differs in how the curriculum is constructed. Rather than repeatedly sampling from a fixed collection, we generate new variants from problems that the model can solve often but not consistently. These variants become the basis for later rounds, allowing the training distribution to change as the model improves.

\paragraph{Synthetic Data and Self Evolving.}
Synthetic data is widely used to expand reasoning datasets when human written problems and solutions are limited. Early work generates mathematical training data through question rewriting, iterative question composition, and program based transformations \citep{yu2024metamathbootstrapmathematicalquestions,liu2024augmentingmathwordproblems,khan2025executablefunctionalabstractionsinferring}. More recent methods place generation inside the training process, using the model's weaknesses, current ability, or previous outputs to produce new problems for later training \citep{chen2025selfevolvingcurriculumllmreasoning,liang2025pass1selfplayvariationalproblem,liang2025swsselfawareweaknessdrivenproblem,zhang2026d2evodualdifficultyawareselfevolution,röpke2026dejaqopenendedevolutiondiverse,huang2026rzeroselfevolvingreasoningllm}. At the same time, repeated training on model generated data can reduce diversity or cause performance to regress \citep{luo2026learningsyntheticdatamodel,lin2026selfimprovementselfregressriseandcollapsefailure}. Our method addresses both concerns by generating variants from problems that match the model's current ability and allowing successful variants to seed later rounds. This keeps the data learnable while changing its structure across rounds, so the training distribution develops together with the model rather than being generated once at the start.

\paragraph{Learning Ceilings in Reinforcement Fine Tuning.}
Prior work offers different explanations for why reinforcement learning with verifiable rewards eventually stops improving. Some work connects this plateau to declining policy entropy and the resulting loss of exploration during training \citep{cui2025entropymechanismreinforcementlearning}. A separate line of work focuses on the training data, showing that problems provide little useful signal once they become too easy, while problems that are too difficult rarely produce successful samples \citep{bae2026onlinedifficultyfilteringreasoning,huang2026emergenceimplicitcurriculumrlvr}. Recent results suggest that these limits can be overcome through longer training, better control of diversity, or curricula that introduce useful reasoning patterns near the model's capability boundary \citep{liu2025prorlprolongedreinforcementlearning,yuan2026understandingdiversitycollapserlvr,cai2026curriculumreinforcementlearningincentivize}. Our work studies the role of the training distribution in this debate. Under the same data budget, static training reaches a plateau, while repeatedly generating variants from problems suited to the model's current ability continues to improve performance.

%% file: sections/method.tex
\section{Method}
\label{sec:method}

Our method couples two components. Question-begets-Question (QbQ) is a synthetic data generation procedure in which a teacher model turns an existing problem into new problems that probe a slightly different aspect of the same skills required to solve the existing problem. The self-evolving curriculum decides which problems QbQ expands and what the model trains on.
Each of $T$ rounds of the curriculum runs the following three stages: (1)Self-Evaluation: the current checkpoint is evaluated on the previous round's training problems and categorizes them into four bands, the problems it already mastered, the problems it mostly get right, the problems it sometimes get right, and the problems it never solved; (2) QbQ Generation: the band of problems it can mostly get right become seeds, and QbQ expands exactly those into fresh variants; (3)Training: we apply GRPO on the new variants, using only their problem statements and answers generated by the teacher model. No step-by-step solution is used for training. This setup both mirrors settings in which oracle reasoning traces are unavailable and mitigates the concern that the gains are distilled from the teacher.
Algorithm~\ref{alg:loop} details the QbQ self-evolving curriculum.

\subsection{Setup and Notation}
\label{sec:setup}

We formalize a problem to be a pair $p=(x_p,a_p)$: a problem statement $x_p$ and an answer.
A policy $\pi$ maps $x_p$ to a sampled solution $y$, and $\mathrm{ans}(y)$ denotes the final answer.

For each problem $p$, to determine which difficulty band it belongs to with respect to the model checkpoint at round $t$, which is $\pi_t$, we draw $k=16$ i.i.d.\ samples $y^{(1)},\dots,y^{(k)}\sim\pi_t(\cdot\mid x_p)$ under fixed decoding parameters and count the number of correct final answers, denoted as $n$.

Self-evaluation sorts problems into four bands by this count: \emph{mastered} ($n=16$), problems the model can \emph{mostly} get right ($8\le n\le 15$), problems it \emph{sometimes} gets right ($1\le n\le 7$), and problems it \emph{never} gets right ($n=0$). The band where the seed questions are is denoted as the seed band $B=\{\,n:k/2\le n\le k{-}1\,\}$, which means $B= \{8\le n\le 15\}$

For our QbQ synthetic data generation procedure, we use problems from the second band as seeds. Problems in this band has the following characteristics: the model checkpoint $\pi_t$'s comprehension of these problems is incomplete, yet success is frequent enough that reinforcement learning algorithms have enough positive signals to learn from. Whereas for problems with $n=0$ and problems with $n=16$, both give every rollout in a group approximately the same reward and hence \emph{zero} gradient (formalized in \S\ref{sec:train}).

Generating questions from this band of ($8 \leq n \leq 15$) is a deliberate design choice, and our experiments will show that this design empirically outperforms the other design which is when we generate questions from the band ($1 \leq n \leq 7$). See Appendix \ref{app:additional_exp} for details.

\paragraph{Train/Test Split}
The training pool $\mathcal{D}$ contains 1{,}005 AIME problems (1983--2024); the two most recent competition years are held out entirely for testing.

We use Qwen2.5-Math-7B for our main experiments, since it is released in 2024 and does not present any risk of contamination for our test set. We obtain the round-zero checkpoint $\pi_0$ by one supervised fine-tuning pass over teacher-written solutions to the pool via LoRA \citep{hu2021loralowrankadaptationlarge}. This SFT step is done to ensure Qwen2.5-Math-7B understands the correct format for answering questions. Empirically the official solution is too abstract for Qwen2.5-Math-7B to understand, so we use GPT5-mini to elaborate on the solution. Other than the initialization step, GPT5-mini's reasoning traces or solutions are never used for training our model.

The initial set of problems is the mostly-right band of $\pi_0$ over the pool, which yields $133$ seeds for our main backbone.

\begin{algorithm}[t]
\caption{The QbQ self-evolving curriculum}
\label{alg:loop}
\begin{algorithmic}[1]
\REQUIRE pool $\mathcal{D}$; initial checkpoint $\pi_0$; rounds $T$; rollout budget $k$;
         seed band $B=\{\,n:k/2\le n\le k{-}1\,\}$
\STATE $S_0 \gets \{\,p\in\mathcal{D} \,:\, n_0(p)\in B\,\}$ \hfill$\triangleright$ mostly-right band of $\pi_0$ over the pool
\FOR{$t=0,\dots,T-1$}
  \STATE $V_t \gets \textsc{QbQ}(S_t)$ \hfill$\triangleright$ synthesize and filter variants (\S\ref{sec:synth})
  \STATE score each $q\in V_t$: $k$ rollouts of $\pi_t \Rightarrow n_t(q)$
  \STATE $R_t \gets \textsc{SelectRL}(V_t)$
  \STATE $\pi_{t+1} \gets \textsc{GRPO}(\pi_t,\,R_t)$
  \STATE re-score each $q\in R_t$: $k$ rollouts of $\pi_{t+1} \Rightarrow n_{t+1}(q)$
  \STATE $S_{t+1} \gets \{\,q\in R_t \,:\, n_{t+1}(q)\in B\,\}$
\ENDFOR
\RETURN $\pi_T$
\end{algorithmic}
\end{algorithm}

\subsection{Question-begets-Question}
\label{sec:synth}

At each round, QbQ transforms the current seeds $S_t$ into a set of new training problems $V_t$ (Algorithm~\ref{alg:loop}, line~3).
An unconstrained prompt such as ``write a similar problem'' tends to produce either superficial rewrites that preserve the underlying computation or unrelated problems that no longer test the target skill.
The former yields near-duplicate training examples, whereas the latter introduces uncontrolled skill drift.
To avoid these failure modes, QbQ conditions the teacher on one of five human-designed \textbf{structural transformation operators}.
Each operator changes the problem structure while preserving its core skill, allowing the resulting variants to probe complementary aspects of that skill without systematically increasing difficulty.

\begin{itemize}
\item \textbf{Generalize, then specialize.}
Replace a fixed quantity in the seed with a parameter, then instantiate it in a new regime where the same lemma applies differently.
For example, ``trailing zeros of $100!$'' in base $10$ becomes ``trailing zeros of $100!$ in base~$12$,'' which requires considering the exponents of both $2$ and $3$.
\item \textbf{Parametrize and sum.}
Introduce an index over configurations related to the seed and ask for an aggregate, with each term obtained using the same lemma.
For example, ``ordered sums of $4$ positive integers equal to $20$'' becomes ``$\sum_{j=1}^{5} f(j)$, where $f(j)$ counts ordered sums of $j$ positive integers equal to $20$.''
\item \textbf{Change the queried quantity.}
Retain the setup and givens but query a different quantity that still requires the seed's key lemma.
For example, for a triangle with side lengths $13$, $14$, and $15$, querying the inradius instead of the area still requires Heron's formula.
\item \textbf{Inverse problem.}
Fix the value of the original target and solve in the reverse direction for a parameter that attains it.
For example, ``how many positive solutions does $20m+12n=2012$ have?'' becomes ``find the least $c$ for which $20m+12n=c$ has exactly $5$.''
\item \textbf{Add a constraint layer.}
Embed the seed computation within one additional routine reduction, without requiring a new theorem.
For example, ``$7^{2024} \bmod 100$'' becomes ``$(7^{2024}+3^{2024}) \bmod 100$.''
\end{itemize}

We additionally require every variant to have an answer different from that of its parent.
This inexpensive check rejects exact copies and answer-preserving surface rewrites, complementing the structural constraints imposed by the operators.

\paragraph{Planning and generation.}
The applicability of an operator depends on the seed.
For example, a geometry problem may not admit a natural parametrization-and-summation transformation, while a fixed numerical computation may not have a meaningful inverse.
For each seed, a planning call receives all five operator descriptions and selects the three most suitable operators.
Invalid selections are replaced in a fixed order.
This planning step encourages the variants of each seed to follow distinct structural directions.

Each selected (seed, operator) pair is then processed by a separate generation call.
The teacher receives the seed, a reference solution, and the complete operator instruction.
Changing only numerical values is explicitly disallowed, and the teacher may abstain if the operator is not applicable.
Only the statement and answer are retained for training, consistent with the protocol in \S\ref{sec:train}.

\paragraph{Diversity of the generated pool.}
We quantify problem diversity using the Vendi score, defined as the exponentiated entropy of the eigenspectrum of the pairwise cosine-similarity matrix over Qwen3-Embedding-8B representations.
The score can be interpreted as the effective number of distinct problems in a pool.
We compare QbQ synthetic data with standard synthetic data generated from real AIME training questions. After normalization, we show that QbQ achieves a higher within-family Vendi score than standard synthetic data generation ($1.59$ vs.\ $1.25$).


\subsection{Self-Training}
\label{sec:train}
For each variant $q\in V_t$, we estimate its pre-update solve count $n_t(q)$, its number of correct answers among the $k$ rollouts drawn during self-evaluation under $\pi_t$ (Algorithm~\ref{alg:loop}, line~4), following the definition in \S\ref{sec:setup}.
These scores are used directly to construct the RL set, so selection requires no additional rollouts.
\paragraph{Selecting RL problems (\textsc{SelectRL}).}
The GRPO learning signal for a problem is determined by reward variation within its sampled group.
For a binary correctness reward with empirical solve rate $\hat p=n_t(q)/k$, the within-group standard deviation is $\smash{\sqrt{\hat p(1-\hat p)}}$.
It is zero at $\hat p\in\{0,1\}$ and maximized at $\hat p=\tfrac12$, corresponding to $n_t(q)=k/2$ under our difficulty estimate.
We therefore construct the RL set $R_t$ of $N=300$ problems from the eligible variants
$\{q\in V_t:1\le n_t(q)\le k{-}1\}$ in two stages.
First, to preserve parent coverage, we select for each seed its eligible variant whose solve count is closest to $k/2$.
We then fill the remaining slots in increasing order of $|n_t(q)-k/2|$, subject to a per-seed cap.

\paragraph{GRPO.}
For each $q\in R_t$, we sample $m=8$ completions
$y_1,\dots,y_m\sim\pi_{\mathrm{old}}(\cdot\mid x_q)$, where $\pi_{\mathrm{old}}$ is anchored at $\pi_t$ at the start of the round, and compute a reward and group-normalized advantage for each completion:
\begin{equation}
\label{eq:reward}
\begin{aligned}
r_i&=\mathbf{1}\bigl[\mathrm{ans}(y_i)=a_q\bigr]+0.1\,b_i,\\
\hat A_i&=(r_i-\bar r)\,/\,s_r,
\end{aligned}
\end{equation}
where $b_i\in\{0,1\}$ indicates compliance with the prescribed answer format, and the small bonus encourages well-formed completions.
The quantities $\bar r$ and $s_r$ denote the mean and standard deviation of $r_{1:m}$.
Let $\rho_{i,\tau}(\theta)$ be the per-token importance ratio with respect to $\pi_{\mathrm{old}}$.
We maximize the clipped GRPO objective \citep{shao2024deepseekmathpushinglimitsmathematical}
\begin{equation}
\label{eq:grpo}
\begin{gathered}
\bar\rho_{i,\tau}=\operatorname{clip}\bigl(\rho_{i,\tau},\,1{-}\varepsilon,\,1{+}\varepsilon\bigr),\\
\ell_{i,\tau}=\min\bigl(\rho_{i,\tau}\hat A_i,\ \bar\rho_{i,\tau}\hat A_i\bigr),\\
\mathcal{J}(\theta)=\mathbb{E}\Bigl[\frac{1}{m}\sum_{i=1}^{m}\frac{1}{|y_i|}\sum_{\tau}\ell_{i,\tau}\Bigr],
\end{gathered}
\end{equation}
with clipping parameter $\varepsilon=0.2$ and no KL penalty.
We limit each round to $150$ optimization steps, corresponding to one pass over $R_t$, and refresh the training set in every round.
The resulting policy is denoted by $\pi_{t+1}$.

\subsection{Evolving the Curriculum}
\label{sec:evolve}
Let $B=\{\,n:k/2\le n\le k-1\,\}$ denote the \emph{mostly-right} band from \S\ref{sec:setup} (with $k=16$, this is $8\le n\le 15$).
After the round-$t$ update produces $\pi_{t+1}$, we re-evaluate the problems in $R_t$ under $\pi_{t+1}$, obtaining updated solve counts $n_{t+1}(q)$, and retain those that fall in the seed band:
\begin{equation}
\label{eq:seed}
S_{t+1}\;=\;\bigl\{\,q\in R_t \;:\; n_{t+1}(q)\in B\,\bigr\}.
\end{equation}
Here $S_{t+1}$ is the seed set that QbQ expands in round $t{+}1$.
Problems mastered after training, with $n_{t+1}(q)=k$, leave the curriculum.
Problems that remain in $B$ become seeds for the next round, while those below the band ($n_{t+1}(q)\le k/2-1$) are discarded without replacement.
Because $R_t$ contains variants with solve counts from $1$ to $k{-}1$, training can promote an initially below-band problem into $B$: such a problem enters the curriculum only once it becomes learnable for the current policy.
The training distribution therefore evolves with model competence rather than following a predefined difficulty schedule.
The number of rounds $T$ is fixed in advance, and held-out performance is never used for selection or stopping.
Beyond initialization, the only external supervision used by the loop consists of the pool and the synthetic problem answers.

%% file: sections/experiment.tex
%
%

\section{Experiments}
\label{sec:exp}

We study the three obstacles identified in Section \ref{sec:introduction} in a
controlled setting where AIME is difficult for the base model and real
training data are limited. Our experiments ask: (1) how far can
standard training go on the available real problems; (2) does QbQ
produce more useful synthetic data than static augmentation; and (3)
with the QbQ data and training budget fixed, does organizing the data
as a curriculum avoid the plateau of non-curriculum training?

\subsection{Experimental Setup}
\label{sec:exp-setup}

\paragraph{Data and model.}
The real-data pool contains $1{,}005$ AIME problems from 1983--2024,
each paired with its official answer. We evaluate on all $60$ problems
from AIME~2025 and AIME~2026. These contests postdate the training pool
and are not used for training, synthesis, curriculum construction, or
model selection. The backbone is Qwen2.5-Math-7B with a
$4{,}096$-token context, and the teacher is
\texttt{gpt-5-mini-2025-08-07}.

\paragraph{Initialization and baselines.}
We first train a supervised initialization, M0, on teacher-written
solutions to the real-data pool. After filtering by answer format and
length, $963$ of the $1{,}005$ traces remain. We train a rank-$32$ LoRA
adapter for three epochs with scale $64$, learning rate $10^{-5}$, and
a cosine schedule. Starting from M0, we apply GRPO for $500$ updates on
the $1{,}005$ original AIME problems.

The static-augmentation baseline continues from the checkpoint after
this GRPO stage. For each original AIME problem, the teacher is asked
for five new problems using the same or closely related techniques. One pass over these variants requires $2{,}443$
additional updates, or $2{,}943$ updates including the first GRPO
stage. This is $57$ updates fewer than the $3{,}000$-update budget used
by the QbQ experiments.

\paragraph{QbQ comparisons.}
QbQ produces $300$ problems in each of $20$ rounds, for a total of
$6{,}000$. We compare two training strategies on exactly the same
problem multiset:
\begin{itemize}
    \item \textbf{QbQ non curriculum} globally shuffles the
    $6{,}000$ problems and trains on them in one GRPO pass.
    \item \textbf{QbQ curriculum} retains the original sequence of
    twenty $300$-problem batches and trains on them in that order.
\end{itemize}
Both strategies use the same initialization, optimizer, rollouts, and
$3{,}000$ GRPO updates. The controlled difference is therefore the
curriculum organization, not the data, compute, or number of updates.
All QbQ RL stages use only problem statements and final answers as
supervision.

\paragraph{GRPO and evaluation.}
GRPO uses groups of eight completions, learning rate
$3\times10^{-6}$, clipping parameter $0.2$, and no KL penalty. The
reward is exact answer correctness plus a $0.1$ answer-format bonus.
Training completions use temperature $1.0$ and a $3{,}072$-token limit;
length-truncated completions are masked from the loss.

At evaluation time, we draw $16$ completions for each held-out problem
with temperature $0.7$, top-$p=0.8$, and a $3{,}072$-token limit.
Pass@1 is the fraction of correct answers across the resulting
$60\times16=960$ completions. We evaluate every final checkpoint with
three sampling seeds and report the mean and standard
deviation.

\subsection{Main Results}
\label{sec:exp-main-results}

\begin{table*}[t]
\centering
\normalsize
\renewcommand{\arraystretch}{1.20}

\begin{tabular*}{\textwidth}{
    @{\extracolsep{\fill}}
    l
    c
    c
    c
    c
    @{}
}
\toprule
\textbf{Model / Training Strategy}
    & \textbf{Seed 0}
    & \textbf{Seed 1}
    & \textbf{Seed 2}
    & \textbf{Mean $\boldsymbol{\pm\sigma}$} \\
\midrule

Qwen2.5-Math-7B (base)
    & 5.83
    & 5.52
    & 5.52
    & $5.62 \pm 0.15$ \\

M0 (supervised initialization)
    & 9.90
    & 9.48
    & 9.06
    & $9.48 \pm 0.34$ \\

M0 + GRPO on original AIME problems
    & 12.08
    & 11.25
    & 11.25
    & $11.53 \pm 0.39$ \\

\quad + Static augmentation
    & 11.98
    & 11.67
    & 12.50
    & $12.05 \pm 0.34$ \\

\midrule

QbQ without curriculum
    & 15.31
    & 14.06
    & 13.96
    & $14.44 \pm 0.61$ \\

\textbf{QbQ curriculum (ours)}
    & \textbf{16.04}
    & \textbf{16.25}
    & \textbf{17.08}
    & $\mathbf{16.46 \pm 0.45}$ \\

\bottomrule
\end{tabular*}

\caption{Pass@1 (\%) averaged over AIME~2025 and AIME~2026. Each
row reports three evaluations of a fixed checkpoint. Both QbQ
variants use the same $6{,}000$ generated problems and $3{,}000$
GRPO updates and differ only in whether the training problems are
organized as a curriculum.}
\label{tab:main-results}
\end{table*}

\paragraph{What each comparison tests.}
The first three rows measure how much can be learned from the original
problem pool: supervised initialization is followed by GRPO on the same
$1{,}005$ problems. Static augmentation then asks whether a standard
one-shot expansion of those problems moves performance beyond that
point. QbQ non curriculum tests whether changing how the synthetic
problems are generated produces a more useful training set. Finally,
the comparison between the two QbQ rows is the strictest ablation in
the table. It holds the $6{,}000$ problems, initialization, optimizer,
and number of updates fixed, and changes only whether the round order
is retained.

\paragraph{Conventional training quickly exhausts the original problem pool.}
The base model reaches $5.62\%$ pass@1, confirming that AIME is well
beyond its initial capability. Supervised initialization improves
pass@1 by $3.86$ points to $9.48\%$, and GRPO on the original AIME
problems reaches $11.53\%$. Further updates on these problems do not
yield a stable gain: in a single-seed checkpoint sweep, pass@1 peaks at
$13.65\%$ after $200$ updates but falls to $12.08\%$ at update $500$.
The three-seed result at the final checkpoint is $11.53\pm0.39\%$.
Because this run contains only $500$ updates, it is a data-limited
baseline rather than a compute-matched comparison with the QbQ runs.

\paragraph{More static data do not remove the plateau.}
Static augmentation increases the training pool by $5{,}000$ usable
variants and nearly matches the QbQ update budget, yet its final
performance is $12.05\pm0.34\%$. This is only $0.52$ points above GRPO
on the original AIME problems. By contrast, QbQ non curriculum reaches
$14.44\pm0.61\%$, improving by $2.39$ points over static augmentation.
Thus QbQ data are more useful than one-shot variants even without a
curriculum, consistent with the Introduction's data-diversity
motivation.

\begin{figure}[t]
\centering
\includegraphics[width=\columnwidth]{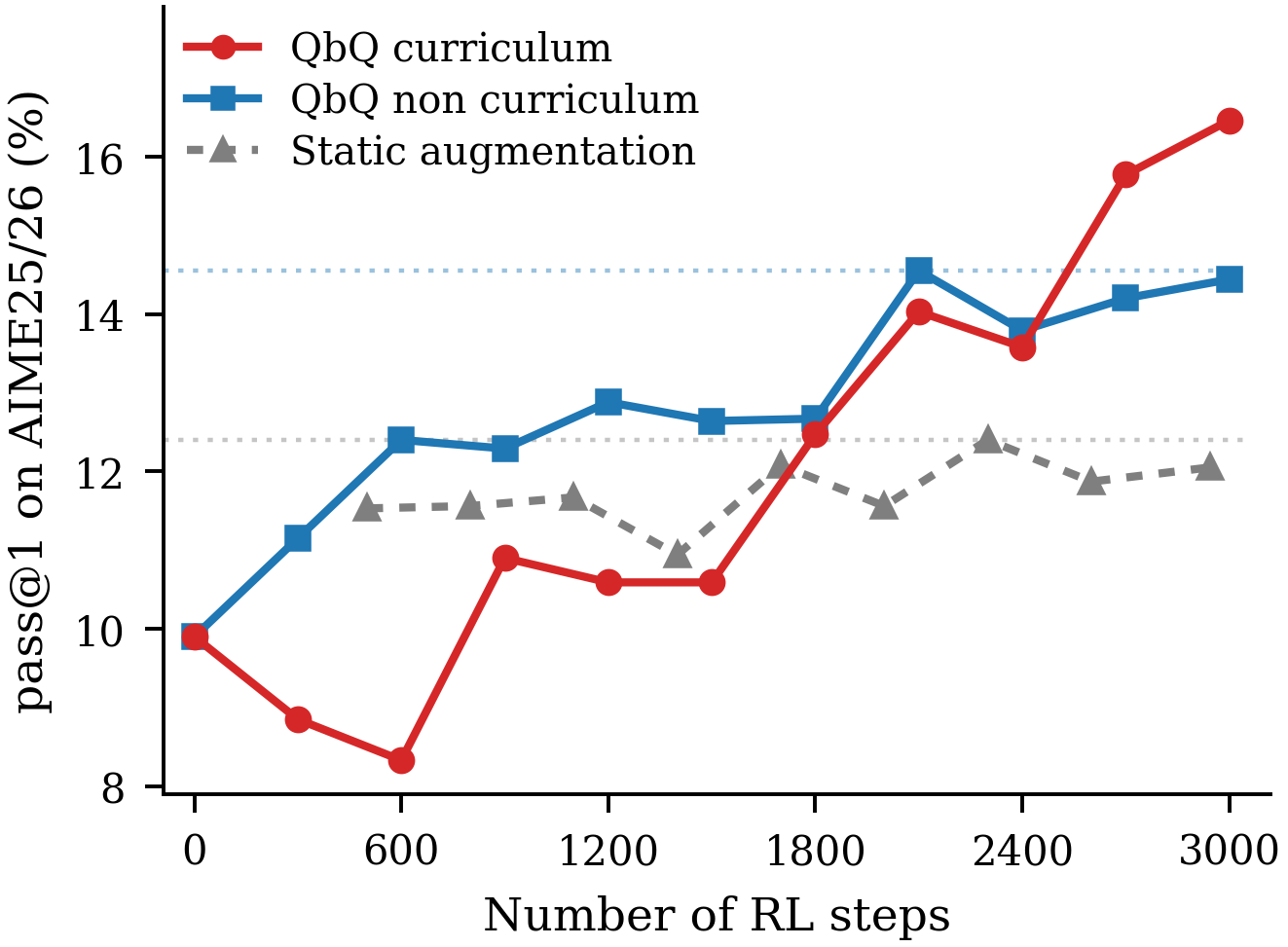}
\caption{Pass@1 on AIME 2025 and 2026 during RL training. Static augmentation and QbQ non curriculum plateau, whereas QbQ curriculum continues to improve through 3,000 updates. Table~\ref{tab:main-results} reports three-seed averages at the final checkpoints.}
\label{fig:rl-curves}
\end{figure}

\paragraph{The curriculum changes the learning trajectory.}
Figure~\ref{fig:rl-curves} shows why endpoint accuracy alone understates
the difference. QbQ non curriculum reaches $14.55\%$ at $2{,}100$
updates, then records $13.79\%$, $14.20\%$, and $13.96\%$ at
$2{,}400$, $2{,}700$, and $3{,}000$ updates. The last $30\%$ of
training therefore fails to improve on the earlier checkpoint.
Static augmentation is more limited still: its checkpoint evaluations
never exceed $12.40\%$ and show no sustained upward trend. Its
three-seed endpoint in Table~\ref{tab:main-results} is
$12.05\pm0.34\%$. Over the same run, its training reward rises from
approximately $0.28$ to $0.54$. The model is therefore fitting the
fixed training pool more successfully without a corresponding gain on
the held-out contests.

QbQ non curriculum exhibits the same problem at a higher accuracy.
Each of its $6{,}000$ problems is visited only once, so the plateau is
not caused by training for multiple epochs on the same examples. The
final $30\%$ of training consumes $1{,}800$ additional QbQ problems
but does not exceed the result at $2{,}100$ updates. Both fixed
training distributions thus reach an observed ceiling within the
available budget.

QbQ curriculum behaves differently. In the plotted trajectory, pass@1
rises from $13.58\%$ at $2{,}400$ updates to $15.77\%$ at $2{,}700$
and $17.08\%$ at $3{,}000$. It gains $3.50$ points over the last
$600$ updates and finishes at its best plotted checkpoint, with no
observed ceiling. The three-seed endpoint is
$16.46\pm0.45\%$, $2.02$ points above QbQ non curriculum.

Because the two QbQ strategies use the same examples and compute, this
controlled comparison attributes the different trajectories to
curriculum organization. The shape of the curves is consistent with
the intended mechanism: global shuffling mixes problems produced for
different stages of model competence, whereas the curriculum retains
the round-by-round progression. Within the tested budget, this ordering
separates the plateau from continued improvement.

\paragraph{Near-successes appear to provide better seeds than harder problems.}
We also vary the difficulty of the seed set. The main curriculum begins from 133 problems that M0 solves often but not reliably; a control run begins from 133 problems that M0 sometimes gets right. More details are provided in Appendix \ref{app:additional_exp}.

\subsection{Transfer Beyond the Curriculum}
\label{sec:exp-zero-solve}

We next test whether curriculum training transfers to problems that it
does not select for RL. We define a fixed hard subset of $499$ real
problems on which M0 has zero sampled accuracy. These problems are
outside the curriculum's seed band and are never selected as seeds.
After QbQ curriculum training, aggregate pass@1 on this subset rises
from $0.0\%$ to $5.3\%$.

%% file: sections/conclusion.tex

\section{Conclusion}
\label{sec:conclusion}

We introduce Question-begets-Question (QbQ), a self-evolving curriculum that teaches a
model a skill it has not mastered without any ground-truth reasoning traces. QbQ lifts
Qwen2.5-Math-7B from $5.62\%$ to $16.46\%$ pass@1 on AIME 2025 and 2026 and keeps
improving through $3{,}000$ GRPO updates with no observed ceiling, while static
augmentation and shuffled synthetic data both plateau under the same budget. Trained with
our novel operator-based synthesis, QbQ generates markedly more diverse problems than
one-shot augmentation ($1.59$ versus $1.25$ Vendi score) and outperforms it by $4.41$
points; holding the $6{,}000$ problems, initialization, optimizer, and update count fixed
and changing only the round ordering still yields $2.02$ points, isolating curriculum
organization as the source of the gain. By reseeding each round with problems that the current checkpoint solves often but not reliably, QbQ keeps rollout groups within GRPO’s high variance regime. It also outperforms seeding from the model’s hardest failures, achieving (16.46\%) compared with (11.36\%), contrary to the recommendation of mistake driven synthesis.
The gains
generalize beyond the curriculum's own distribution: on $499$ real problems the model never
solves and the loop never selects as seeds, pass@1 rises from $0.0\%$ to $5.3\%$. Because the policy trains only on problem statements and verified final answers, never on teacher reasoning, QbQ shows that an apparent learning ceiling may arise from the training distribution rather than the model itself. It also shows that this ceiling can be removed without distillation.

%% file: sections/appendix.tex
%

\section{Additional Experiments}
\label{app:additional_exp}

Our proposed curriculum deliberately generates new problems from the
mostly-right band, $8\le n\le15$, which contains problems the model solves
often but not yet reliably (\S\ref{sec:synth}). We compare it with two
intuitive alternatives. The harder-seed curriculum uses seeds with solve
counts $1\le n\le7$, whereas the mixed-seed curriculum uses seeds with solve
counts $1\le n\le15$, where $n$ denotes the number of correct completions
among $16$ rollouts. We rerun the complete self-evolving loop with each seed
band. Neither alternative reproduces the gain of the mostly-right curriculum.

\paragraph{Setup.}
The comparison uses the same initialization, teacher, operator library,
generation framework, and held-out evaluation as the main run
(Table~\ref{tab:hyper}). Each arm is configured for $20$ rounds, with $300$ RL
problems and $150$ optimizer steps per round. The arms use the following seed
bands:
\begin{itemize}\setlength\itemsep{2pt}
\item \textbf{Harder seeds ($1\le n\le 7$).}
The initial set contains $133$ problems sampled from the $362$ problems that
M0 solves between $1$ and $7$ times in $16$ attempts. This matches the size
of the main run's initial seed set.
\item \textbf{Mixed seeds ($1\le n\le 15$).}
The initial set contains $133$ problems sampled from the union of the two
nondegenerate bands ($495$ problems). Stratification by solve count gives
$96$ sometimes-right and $37$ mostly-right seeds, approximately matching the
composition of the full pool.
\end{itemize}

\begin{table}[hbpt]
\centering
\small
\setlength{\tabcolsep}{3pt}
\renewcommand{\arraystretch}{1.0}
\begin{tabular}{@{}lcccc@{}}
\toprule
\textbf{Training condition}
    & \textbf{$s=0$} & \textbf{$s=1$} & \textbf{$s=2$}
    & \textbf{Mean $\boldsymbol{\pm\sigma}$} \\
\midrule
Original-problem GRPO
    & 12.08 & 11.25 & 11.25 & $11.53 \pm 0.39$ \\
\midrule
\textbf{Mostly right ($8$--$15$)}
    & \textbf{16.04} & \textbf{16.25} & \textbf{17.08}
    & $\mathbf{16.46 \pm 0.45}$ \\
Harder ($1$--$7$)
    & 11.67 & 11.15 & 11.25 & $11.35 \pm 0.23$ \\
Mixed ($1$--$15$)
    & 12.29 & 11.88 & 11.56 & $11.91 \pm 0.30$ \\
\bottomrule
\end{tabular}
\caption{Endpoint Pass@1 (\%) on AIME~2025 and AIME~2026. Each entry averages
the two contest years over three evaluation seeds. The original-problem GRPO
result is reproduced from Table~\ref{tab:main-results} as a reference; the
remaining rows compare curricula defined by different seed bands. $\sigma$
denotes the population standard deviation across evaluation seeds.}
\label{tab:seed-bands}
\end{table}

\begin{figure}[t]
\centering
\includegraphics[width=\columnwidth]{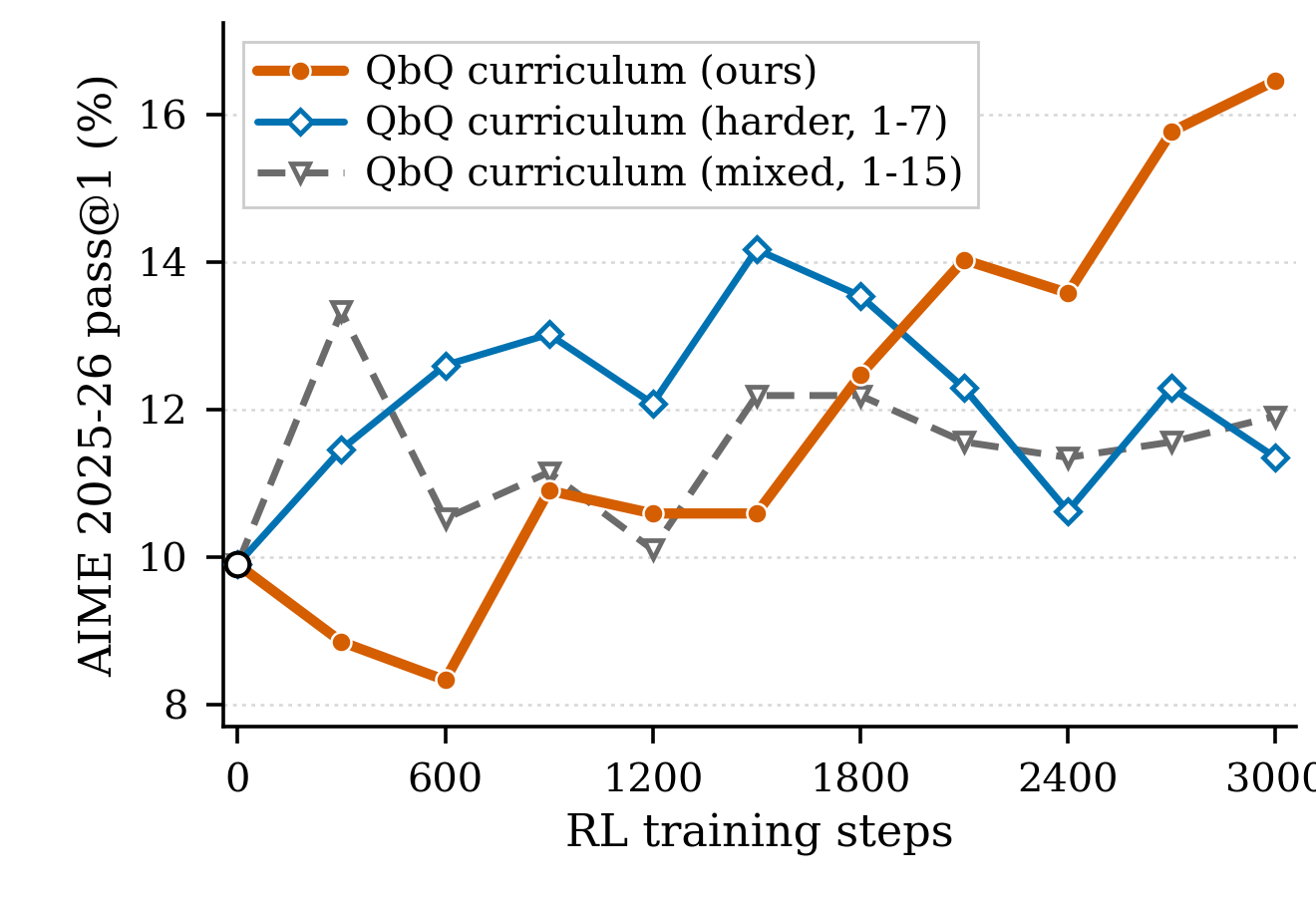}
\caption{Held-out Pass@1 during training for the three seed bands.
Intermediate points use one evaluation seed; each endpoint is the three-seed
mean in Table~\ref{tab:seed-bands}. The mostly-right curriculum is the only
arm that continues improving through the end of training. The harder-seed
curriculum peaks midway and then declines, whereas the mixed curriculum never
escapes its early plateau.}
\label{fig:seed-bands}
\end{figure}

\paragraph{Results.}
Table~\ref{tab:seed-bands} and Figure~\ref{fig:seed-bands} show the outcome.
The harder- and mixed-seed curricula finish at $11.35\pm0.23\%$ and
$11.91\pm0.30\%$, respectively. Both endpoints lie within $0.4$ percentage
points of the original-problem GRPO baseline and fail to reproduce the gain of
the mostly-right curriculum, which reaches $16.46\pm0.45\%$. The seed band is
consequently not a minor tuning detail: it separates a curriculum that works
from curricula that contribute no material gain.

The two alternatives fail in different ways. The harder-seed curriculum
initially outpaces the main run and reaches $14.17\%$ halfway through
training, but then loses the entire gain. The mixed curriculum peaks early at
$13.33\%$ and subsequently fluctuates between approximately $10\%$ and
$12.2\%$, never escaping its initial plateau. The mostly-right curriculum
follows the opposite trajectory: it trails both alternatives early, improves
sharply late in training, and finishes at its maximum observed value.

\paragraph{Why harder seeds fail.}
A harder seed does not necessarily provide a stronger learning signal. Its
variants are more likely to remain beyond the model's reliable solving range,
producing rollout groups with no correctness variation. When the format reward
is also constant, their standardized advantages vanish; even otherwise, they
provide no relative correctness signal. By contrast, variants of mostly-right
parents tend to lie near the model's competence boundary, where groups contain
both correct and incorrect completions and yield informative relative
advantages. The harder majority therefore shifts the mixed curriculum toward
lower-signal groups.

These results support the paper's central causal claim: progress comes from
generating just beyond the model's current competence boundary, not from
maximizing seed difficulty. Problems that are already solved often but not
reliably provide the productive frontier for self-evolution; shifting the
curriculum toward harder seeds removes that advantage.

\section{The Operator Details}
\label{app:ops}

This section provides the complete definitions and worked examples for the
five structural transformation operators introduced in \S\ref{sec:synth}.
Each operator changes the problem structure while preserving its core
mathematical skill. The examples are deliberately elementary so that the
structural transformation is transparent; they are illustrative rather than
training instances. Every generated variant is additionally required to have
an answer different from that of its parent.

\subsection{Operator Definitions and Worked Examples}

\paragraph{O1: Generalize, then specialize.}
Identify a value that the seed holds fixed, promote it to a free parameter,
and specialize the parameter to a different concrete case in which the same
lemma applies in a new regime. Merely increasing a numerical value does not
qualify.

\begin{itemize}\setlength\itemsep{1pt}
\item[] \textbf{Seed.} Find the number of trailing zeros of $100!$ when
written in base $10$.
\item[] \textbf{Variant.} Find the number of trailing zeros of $100!$ when
written in base $12$.
\item[] \textbf{Analysis.} Both problems use Legendre's formula for prime
exponents in a factorial. The seed fixes the base as $10=2\cdot5$ and reduces
to counting factors of $5$. The variant uses $12=2^2\cdot3$, so the solver
must compare the exponents of $2$ and $3$. The lemma is unchanged, but it is
applied in a structurally different regime.
\end{itemize}

\paragraph{O2: Parametrize and sum.}
Starting from a seed that concerns one configuration, introduce an index over
a small family of related configurations and ask for an aggregate. Each term
must be obtained using the seed's lemma; the aggregation supplies the new
structure.

\begin{itemize}\setlength\itemsep{1pt}
\item[] \textbf{Seed.} In how many ways can $20$ be written as an ordered sum
of $4$ positive integers?
\item[] \textbf{Variant.} For $j=1,\ldots,3$, let $f(j)$ be the number of ways
to write $20$ as an ordered sum of $j$ positive integers. Find
$\sum_{j=1}^{3}f(j)$.
\item[] \textbf{Analysis.} Each term is a composition count,
$f(j)=\binom{19}{j-1}$, obtained by the same stars-and-bars argument as the
seed. The variant treats this count as a function of $j$ and aggregates its
values over a family of configurations, giving $1+19+171=191$.
\end{itemize}

\paragraph{O3: Change the queried quantity.}
Retain the setup, given values, and governing relations, but ask for a
different unknown that still requires the seed's key lemma.

\begin{itemize}\setlength\itemsep{1pt}
\item[] \textbf{Seed.} A triangle has side lengths $13$, $14$, and $15$.
Find its area.
\item[] \textbf{Variant.} A triangle has side lengths $13$, $14$, and $15$.
Find its inradius.
\item[] \textbf{Analysis.} Both solutions require Heron's formula, which
gives area $84$. The seed stops at the area, whereas the variant uses the
additional routine relation $r=K/s=84/21=4$. Thus, the seed's key lemma
remains necessary.
\end{itemize}

\paragraph{O4: Inverse problem.}
When the seed supplies parameters and asks for a resulting value, instead fix
a target value of that result and solve in reverse for a parameter that
attains it. The governing relation remains the same.

\begin{itemize}\setlength\itemsep{1pt}
\item[] \textbf{Seed.} How many ordered pairs of positive integers $(m,n)$
satisfy $20m+12n=2012$?
\item[] \textbf{Variant.} Find the smallest positive integer $c$ for which
$20m+12n=c$ has exactly $5$ solutions in positive integers $(m,n)$.
\item[] \textbf{Analysis.} Both problems count positive solutions of a linear
Diophantine equation. The seed computes the count for a fixed right-hand side,
whereas the variant fixes the count and solves for the right-hand side.
\end{itemize}

\paragraph{O5: Add a constraint layer.}
Embed the seed's core computation in a slightly larger setup that requires one
additional routine reduction back to the core. The added layer must not
require a new theorem.

\begin{itemize}\setlength\itemsep{1pt}
\item[] \textbf{Seed.} Find the remainder when $7^{2024}$ is divided by
$100$.
\item[] \textbf{Variant.} Let $N=7^{2024}+3^{2024}$. Find the remainder when
$N$ is divided by $100$.
\item[] \textbf{Analysis.} The primary skill, reducing large powers modulo
$100$ via periodicity, is unchanged. The variant adds one application of the
same routine to $3^{2024}$, followed by modular addition.
\end{itemize}

\section{Hyperparameters and Protocols}
\label{app:hyper}
\label{app:hparams}

Table~\ref{tab:hyper} summarizes the main-loop and evaluation settings.
Baseline-specific training settings are described in the main text.

\begin{table}[t]
\centering
\small
\setlength{\tabcolsep}{3.5pt}
\begin{tabular}{@{}p{0.59\columnwidth}p{0.32\columnwidth}@{}}
\toprule
\multicolumn{2}{@{}l}{\textit{Loop constants}}\\
backbone / context length & Qwen2.5-Math-7B / $4{,}096$ \\
rollouts per difficulty measurement $k$ & $16$ \\
seed band $B$ & $8\le n\le15$ \\
variants per seed / maximum waves & $3$ / $3$ \\
RL-set size $N$ & $300$ \\
rounds $T$ (fixed in advance) & $20$ \\
GRPO steps per round / total & $150$ / $3{,}000$ \\
\midrule
\multicolumn{2}{@{}l}{\textit{Difficulty measurement}}\\
temperature / top-$p$ / top-$k$ & $0.7$ / $0.8$ / $20$ \\
repetition penalty / token limit & $1.05$ / $3{,}072$ \\
\midrule
\multicolumn{2}{@{}l}{\textit{Initialization of $\pi_0$}}\\
data & $963$ teacher solutions \\
adapter & LoRA $r=32$, $\alpha=64$ \\
dropout / epochs & $0.05$ / $3$ \\
schedule & lr $10^{-5}$, cosine \\
warmup ratio & $0.03$ \\
\midrule
\multicolumn{2}{@{}l}{\textit{Per-round GRPO}}\\
optimizer / learning rate & AdamW / $3\times10^{-6}$ \\
momenta & $\beta_1{=}0.9$, $\beta_2{=}0.99$ \\
gradient-norm clip & $0.2$ \\
group size $m$ / clip $\varepsilon$ / KL & $8$ / $0.2$ / $0$ \\
reward & correctness $+\,0.1\cdot$format \\
sampling temperature / token limit & $1.0$ / $3{,}072$ \\
optimizer steps / total completions per step & $150$ / $16$ \\
\midrule
\multicolumn{2}{@{}l}{\textit{Held-out evaluation}}\\
completions per problem & $16$ \\
temperature / top-$p$ / top-$k$ & $0.7$ / $0.8$ / disabled \\
repetition penalty / token limit & $1.0$ / $3{,}072$ \\
sampling seeds & $3$ (final), $1$ (curves) \\
\bottomrule
\end{tabular}
\caption{Main experimental hyperparameters. LoRA follows
\citet{hu2021loralowrankadaptationlarge}. The backbone is
Qwen2.5-Math-7B; the loop itself is backbone-agnostic.}
\label{tab:hyper}
\end{table}
\paragraph{Prompting and grading.}
All policy sampling uses the backbone's chat template with the instruction
``Please reason step by step, and put your final answer within
\verb|\boxed{}|'' appended to the problem statement. The last boxed integer
is normalized and compared with the reference answer. A completion with no
extractable boxed integer is counted as incorrect.

\paragraph{Supervised initialization.}
The initial policy $\pi_0$ is trained once, before the iterative loop, on one
teacher-written solution per real-pool problem.  A LoRA adapter is trained on
all linear layers with rank $32$, scale $64$, and dropout $0.05$ for three
epochs. We use a learning rate of $10^{-5}$, a cosine schedule, and a warmup
ratio of $0.03$, and then merge the adapter into the backbone. This is the
only stage that trains on solution text; subsequent stages use problem
statements and final answers only (\S\ref{sec:train}).

\paragraph{Synthesis.}
The teacher is \texttt{gpt-5-mini-2025-08-07}. Each round targets three
variants per seed. A planning call selects three of the five operators, and
one generation call is issued for each selected operator.

\paragraph{Difficulty measurement.}
The solve count $n_t(p)$ is the number of correct answers among $k=16$
rollouts sampled with temperature $0.7$, top-$p$ $0.8$, top-$k$ $20$,
repetition penalty $1.05$, and a $3{,}072$-token limit. Because this count is
a sum of sixteen Bernoulli outcomes, re-measurement can move problems near a
band boundary. Each selection therefore uses the count available at selection
time; later re-scoring produces a new count only for subsequent decisions.

\subsection{RL Problem-Set Construction}
\label{app:rlset}

Algorithm~\ref{alg:selectrl} gives the main-run procedure summarized in
\S\ref{sec:train}. First, every seed with an eligible variant retains the
variant whose solve count is closest to $k/2$. This coverage step prevents
lineages from disappearing solely because other seeds have more variants.
The remaining slots are filled by the same distance criterion, subject to a
per-seed cap. Since
$\sqrt{\hat p(1-\hat p)}$ is maximized at $\hat p=1/2$, this ordering
prioritizes problems expected to provide the strongest within-group reward
variation.

Variants with counts $n\in\{0,k\}$ are excluded from the primary pool because
their rollout groups have constant correctness reward at the selection
snapshot. They are used only as a last-resort top-up when the eligible pool is
too small. The main $20$-round run always had more than $N$ eligible variants,
and the degenerate top-up was never used.

\begin{algorithm}[t]
\caption{SelectRL (one round; $N=300$, $k=16$)}
\label{alg:selectrl}
\begin{algorithmic}[1]
\STATE $E \leftarrow
  \{q\in V_t:1\le n_t(q)\le k-1\}$
\STATE $d(q)\leftarrow |n_t(q)-k/2|$
\IF{$|E|\le N$}
  \STATE $R\leftarrow E$
\ELSE
  \STATE $R\leftarrow\emptyset$
  \FOR{each seed $s$ represented in $E$}
    \STATE add to $R$ the variant of $s$ minimizing $d(q)$
  \ENDFOR
  \STATE $\mathrm{cap}\leftarrow
    \lceil N/\#\text{covered seeds}\rceil$
  \FOR{$q\in E\setminus R$ in increasing $d(q)$}
    \STATE add $q$ if its seed has fewer than
      $\mathrm{cap}$ representatives; stop at $|R|=N$
  \ENDFOR
  \IF{$|R|<N$}
    \STATE relax the cap and fill from $E\setminus R$
      in the same order
  \ENDIF
\ENDIF
\IF{$|R|<N$}
  \STATE top up from $n_t=0$, then $n_t=k$
\ENDIF
\RETURN $R$
\end{algorithmic}
\end{algorithm}

\subsection{GRPO Settings}

In the main run, each round performs $150$ optimizer steps, corresponding to
one pass over $N=300$ selected problems at two problems per step. For each
problem, the
policy samples $m=8$ completions at temperature $1.0$ with a
$3{,}072$-token limit; length-truncated completions are masked from the loss.
The reward in Eq.~\eqref{eq:reward} is answer correctness plus a $0.1$
well-formed-answer bonus and is standardized within each group. We use AdamW
with a constant learning rate of $3\times10^{-6}$, no warmup,
$\beta_1=0.9$, $\beta_2=0.99$, gradient-norm clipping at $0.2$, GRPO
clipping parameter $\varepsilon=0.2$, and no KL penalty. Optimizer state is
re-initialized at the beginning of each round, and training uses bfloat16.

\subsection{Computing Infrastructure}

All experiments ran on a single Linux server with NVIDIA H100 80GB GPUs;
each training or evaluation job uses between two and eight GPUs, with four
the typical allocation. The software stack is PyTorch 2.10 with CUDA 12.8,
TRL 1.4.0 for GRPO with vLLM 0.19.1 generating rollouts in colocated mode,
FSDP2 sharding, and Transformers 5.8.1; exact pinned versions of every
package ship with the code appendix. One curriculum round (synthesis,
$150$ GRPO steps, re-scoring, and held-out evaluation) takes roughly $90$
minutes on four GPUs, so a full $20$-round run completes in about $30$
hours; teacher API usage is approximately $\$7$ per round.

\subsection{Held-Out Evaluation}

Held-out accuracy is measured on all $60$ problems from AIME~2025 and
AIME~2026. We draw $k=16$ completions per problem with temperature $0.7$,
top-$p$ $0.8$, top-$k$ disabled, repetition penalty $1.0$, and a
$3{,}072$-token limit. Pass@1 is the fraction of correct answers among the
resulting $960$ completions. Each final checkpoint in the main results table
is evaluated with three sampling seeds, and the mean and standard deviation
are reported. Per-round curves use one fixed sampling seed.

%% file: main.bbl
\begin{thebibliography}{42}
\providecommand{\natexlab}[1]{#1}

\bibitem[{{Art of Problem Solving}(2024{\natexlab{a}})}]{aops2024aime}
{Art of Problem Solving}. 2024{\natexlab{a}}.
\newblock {AIME} Problems and Solutions.
\newblock Art of Problem Solving Wiki,
  \url{https://artofproblemsolving.com/wiki/index.php/AIME_Problems_and_Solutions}.
\newblock Accessed 2026-07-29.

\bibitem[{{Art of Problem Solving}(2024{\natexlab{b}})}]{aops2024usamo}
{Art of Problem Solving}. 2024{\natexlab{b}}.
\newblock {USAMO} Problems and Solutions.
\newblock Art of Problem Solving Wiki,
  \url{https://artofproblemsolving.com/wiki/index.php/USAMO_Problems_and_Solutions}.
\newblock Accessed 2026-07-29.

\bibitem[{Bae et~al.(2026)Bae, Hong, Lee, Kim, Nam, and
  Kwak}]{bae2026onlinedifficultyfilteringreasoning}
Bae, S.; Hong, J.; Lee, M.~Y.; Kim, H.; Nam, J.; and Kwak, D. 2026.
\newblock Online Difficulty Filtering for Reasoning Oriented Reinforcement
  Learning.
\newblock arXiv:2504.03380.

\bibitem[{Cai et~al.(2026)Cai, Fang, Li, Zeng, Li, and
  Chen}]{cai2026curriculumreinforcementlearningincentivize}
Cai, P.; Fang, T.; Li, X.; Zeng, Q.; Li, G.; and Chen, J. 2026.
\newblock Curriculum Reinforcement Learning Can Incentivize Reasoning Capacity
  in LLMs Beyond the Base Model.
\newblock arXiv:2606.22317.

\bibitem[{Chen et~al.(2026{\natexlab{a}})Chen, Peng, Choubey, Huang, Zhang,
  Bansal, and Wu}]{chen2026nudgingboundariesllmreasoning}
Chen, J. C.-Y.; Peng, B.~X.; Choubey, P.~K.; Huang, K.-H.; Zhang, J.; Bansal,
  M.; and Wu, C.-S. 2026{\natexlab{a}}.
\newblock Nudging the Boundaries of LLM Reasoning.
\newblock arXiv:2509.25666.

\bibitem[{Chen et~al.(2026{\natexlab{b}})Chen, Prasad, Khan, Singh, Tian,
  Stengel-Eskin, and
  Bansal}]{chen2026cogdriftexplorationadaptivelyreformulated}
Chen, J. C.-Y.; Prasad, A.; Khan, Z.; Singh, J.; Tian, R.; Stengel-Eskin, E.;
  and Bansal, M. 2026{\natexlab{b}}.
\newblock Cog-DRIFT: Exploration on Adaptively Reformulated Instances Enables
  Learning from Hard Reasoning Problems.
\newblock arXiv:2604.04767.

\bibitem[{Chen et~al.(2025)Chen, Lu, Kim, Zhang, Tang, Piché, Gontier, Bengio,
  and Kamalloo}]{chen2025selfevolvingcurriculumllmreasoning}
Chen, X.; Lu, J.; Kim, M.; Zhang, D.; Tang, J.; Piché, A.; Gontier, N.;
  Bengio, Y.; and Kamalloo, E. 2025.
\newblock Self-Evolving Curriculum for LLM Reasoning.
\newblock arXiv:2505.14970.

\bibitem[{Cobbe et~al.(2021)Cobbe, Kosaraju, Bavarian, Chen, Jun, Kaiser,
  Plappert, Tworek, Hilton, Nakano, Hesse, and
  Schulman}]{cobbe2021trainingverifierssolvemath}
Cobbe, K.; Kosaraju, V.; Bavarian, M.; Chen, M.; Jun, H.; Kaiser, L.; Plappert,
  M.; Tworek, J.; Hilton, J.; Nakano, R.; Hesse, C.; and Schulman, J. 2021.
\newblock Training Verifiers to Solve Math Word Problems.
\newblock arXiv:2110.14168.

\bibitem[{Cui et~al.(2025)Cui, Zhang, Chen, Yuan, Wang, Zuo, Li, Fan, Chen,
  Chen, Liu, Peng, Bai, Ouyang, Cheng, Zhou, and
  Ding}]{cui2025entropymechanismreinforcementlearning}
Cui, G.; Zhang, Y.; Chen, J.; Yuan, L.; Wang, Z.; Zuo, Y.; Li, H.; Fan, Y.;
  Chen, H.; Chen, W.; Liu, Z.; Peng, H.; Bai, L.; Ouyang, W.; Cheng, Y.; Zhou,
  B.; and Ding, N. 2025.
\newblock The Entropy Mechanism of Reinforcement Learning for Reasoning
  Language Models.
\newblock arXiv:2505.22617.

\bibitem[{Dahl et~al.(2024)Dahl, Magesh, Suzgun, and
  Ho}]{dahl2024largelegalfictionsprofiling}
Dahl, M.; Magesh, V.; Suzgun, M.; and Ho, D.~E. 2024.
\newblock Large Legal Fictions: Profiling Legal Hallucinations in Large
  Language Models.
\newblock arXiv:2401.01301.

\bibitem[{Glazer et~al.(2025)Glazer, Erdil, Besiroglu, Chicharro, Chen,
  Gunning, Olsson, Denain, Ho, de~Oliveira~Santos, Järviniemi, Barnett,
  Sandler, Vrzala, Sevilla, Ren, Pratt, Levine, Barkley, Stewart, Grechuk,
  Grechuk, Enugandla, and
  Wildon}]{glazer2025frontiermathbenchmarkevaluatingadvanced}
Glazer, E.; Erdil, E.; Besiroglu, T.; Chicharro, D.; Chen, E.; Gunning, A.;
  Olsson, C.~F.; Denain, J.-S.; Ho, A.; de~Oliveira~Santos, E.; Järviniemi,
  O.; Barnett, M.; Sandler, R.; Vrzala, M.; Sevilla, J.; Ren, Q.; Pratt, E.;
  Levine, L.; Barkley, G.; Stewart, N.; Grechuk, B.; Grechuk, T.; Enugandla,
  S.~V.; and Wildon, M. 2025.
\newblock FrontierMath: A Benchmark for Evaluating Advanced Mathematical
  Reasoning in AI.
\newblock arXiv:2411.04872.

\bibitem[{Guha et~al.(2023)Guha, Nyarko, Ho, Ré, Chilton, Narayana,
  Chohlas-Wood, Peters, Waldon, Rockmore, Zambrano, Talisman, Hoque, Surani,
  Fagan, Sarfaty, Dickinson, Porat, Hegland, Wu, Nudell, Niklaus, Nay, Choi,
  Tobia, Hagan, Ma, Livermore, Rasumov-Rahe, Holzenberger, Kolt, Henderson,
  Rehaag, Goel, Gao, Williams, Gandhi, Zur, Iyer, and
  Li}]{guha2023legalbenchcollaborativelybuiltbenchmark}
Guha, N.; Nyarko, J.; Ho, D.~E.; Ré, C.; Chilton, A.; Narayana, A.;
  Chohlas-Wood, A.; Peters, A.; Waldon, B.; Rockmore, D.~N.; Zambrano, D.;
  Talisman, D.; Hoque, E.; Surani, F.; Fagan, F.; Sarfaty, G.; Dickinson,
  G.~M.; Porat, H.; Hegland, J.; Wu, J.; Nudell, J.; Niklaus, J.; Nay, J.;
  Choi, J.~H.; Tobia, K.; Hagan, M.; Ma, M.; Livermore, M.; Rasumov-Rahe, N.;
  Holzenberger, N.; Kolt, N.; Henderson, P.; Rehaag, S.; Goel, S.; Gao, S.;
  Williams, S.; Gandhi, S.; Zur, T.; Iyer, V.; and Li, Z. 2023.
\newblock LegalBench: A Collaboratively Built Benchmark for Measuring Legal
  Reasoning in Large Language Models.
\newblock arXiv:2308.11462.

\bibitem[{Guo and Ding(2026)}]{guo2026largermodelsreallywin}
Guo, J.; and Ding, S. 2026.
\newblock Do Larger Models Really Win in Drug Discovery? A Benchmark Assessment
  of Model Scaling in AI-Driven Molecular Property and Activity Prediction.
\newblock arXiv:2604.26498.

\bibitem[{Hendrycks et~al.(2021)Hendrycks, Burns, Kadavath, Arora, Basart,
  Tang, Song, and
  Steinhardt}]{hendrycks2021measuringmathematicalproblemsolving}
Hendrycks, D.; Burns, C.; Kadavath, S.; Arora, A.; Basart, S.; Tang, E.; Song,
  D.; and Steinhardt, J. 2021.
\newblock Measuring Mathematical Problem Solving With the MATH Dataset.
\newblock arXiv:2103.03874.

\bibitem[{Hu et~al.(2021)Hu, Shen, Wallis, Allen-Zhu, Li, Wang, Wang, and
  Chen}]{hu2021loralowrankadaptationlarge}
Hu, E.~J.; Shen, Y.; Wallis, P.; Allen-Zhu, Z.; Li, Y.; Wang, S.; Wang, L.; and
  Chen, W. 2021.
\newblock LoRA: Low-Rank Adaptation of Large Language Models.
\newblock arXiv:2106.09685.

\bibitem[{Huang et~al.(2026{\natexlab{a}})Huang, Yu, Wang, Zhang, Li, Li,
  Huang, Mi, and Yu}]{huang2026rzeroselfevolvingreasoningllm}
Huang, C.; Yu, W.; Wang, X.; Zhang, H.; Li, Z.; Li, R.; Huang, J.; Mi, H.; and
  Yu, D. 2026{\natexlab{a}}.
\newblock R-Zero: Self-Evolving Reasoning LLM from Zero Data.
\newblock arXiv:2508.05004.

\bibitem[{Huang et~al.(2026{\natexlab{b}})Huang, Wen, Chi, Wei, Singh, Liang,
  and Chen}]{huang2026emergenceimplicitcurriculumrlvr}
Huang, Y.; Wen, Z.; Chi, Y.; Wei, Y.; Singh, A.; Liang, Y.; and Chen, Y.
  2026{\natexlab{b}}.
\newblock On the Emergence of Implicit Curriculum in RLVR Learning Dynamics.
\newblock arXiv:2602.14872.

\bibitem[{{IMO Foundation}(2024)}]{imofoundation2024imo}
{IMO Foundation}. 2024.
\newblock International Mathematical Olympiad.
\newblock \url{https://www.imo-official.org/}.
\newblock Accessed 2026-07-29.

\bibitem[{Karger et~al.(2025)Karger, Bastani, Yueh-Han, Jacobs, Halawi, Zhang,
  and Tetlock}]{karger2025forecastbenchdynamicbenchmarkai}
Karger, E.; Bastani, H.; Yueh-Han, C.; Jacobs, Z.; Halawi, D.; Zhang, F.; and
  Tetlock, P.~E. 2025.
\newblock ForecastBench: A Dynamic Benchmark of AI Forecasting Capabilities.
\newblock arXiv:2409.19839.

\bibitem[{Khan et~al.(2025)Khan, Stengel-Eskin, Prasad, Cho, and
  Bansal}]{khan2025executablefunctionalabstractionsinferring}
Khan, Z.; Stengel-Eskin, E.; Prasad, A.; Cho, J.; and Bansal, M. 2025.
\newblock Executable Functional Abstractions: Inferring Generative Programs for
  Advanced Math Problems.
\newblock arXiv:2504.09763.

\bibitem[{Lee et~al.(2026)Lee, Lu, Diao, Kang, Muralidharan, Sapra, Tao,
  Molchanov, Choi, Wang, and Hachiuma}]{lee2026zoneproximalpolicyoptimization}
Lee, B.-K.; Lu, X.; Diao, S.; Kang, M.; Muralidharan, S.; Sapra, K.; Tao, A.;
  Molchanov, P.; Choi, Y.; Wang, Y.-C.~F.; and Hachiuma, R. 2026.
\newblock Zone of Proximal Policy Optimization: Teacher in Prompts, Not
  Gradients.
\newblock arXiv:2606.18216.

\bibitem[{Liang et~al.(2025{\natexlab{a}})Liang, Li, Gong, Shen, Wu, Guo, and
  Chen}]{liang2025pass1selfplayvariationalproblem}
Liang, X.; Li, Z.; Gong, Y.; Shen, Y.; Wu, Y.~N.; Guo, Z.; and Chen, W.
  2025{\natexlab{a}}.
\newblock Beyond Pass@1: Self-Play with Variational Problem Synthesis Sustains
  RLVR.
\newblock arXiv:2508.14029.

\bibitem[{Liang et~al.(2025{\natexlab{b}})Liang, Li, Gong, Wang, Zhang, Shen,
  Wu, and Chen}]{liang2025swsselfawareweaknessdrivenproblem}
Liang, X.; Li, Z.-Z.; Gong, Y.; Wang, Y.; Zhang, H.; Shen, Y.; Wu, Y.~N.; and
  Chen, W. 2025{\natexlab{b}}.
\newblock SwS: Self-aware Weakness-driven Problem Synthesis in Reinforcement
  Learning for LLM Reasoning.
\newblock arXiv:2506.08989.

\bibitem[{Lightman et~al.(2023)Lightman, Kosaraju, Burda, Edwards, Baker, Lee,
  Leike, Schulman, Sutskever, and Cobbe}]{lightman2023letsverifystepstep}
Lightman, H.; Kosaraju, V.; Burda, Y.; Edwards, H.; Baker, B.; Lee, T.; Leike,
  J.; Schulman, J.; Sutskever, I.; and Cobbe, K. 2023.
\newblock Let's Verify Step by Step.
\newblock arXiv:2305.20050.

\bibitem[{Lin(2026)}]{lin2026selfimprovementselfregressriseandcollapsefailure}
Lin, J. 2026.
\newblock Self-Improvement Can Self-Regress: The Rise-and-Collapse Failure Mode
  of LLM Self-Training.
\newblock arXiv:2606.21090.

\bibitem[{Liu et~al.(2024)Liu, Zhang, Luo, and
  Yao}]{liu2024augmentingmathwordproblems}
Liu, H.; Zhang, Y.; Luo, Y.; and Yao, A. C.-C. 2024.
\newblock Augmenting Math Word Problems via Iterative Question Composing.
\newblock arXiv:2401.09003.

\bibitem[{Liu et~al.(2025)Liu, Diao, Lu, Hu, Dong, Choi, Kautz, and
  Dong}]{liu2025prorlprolongedreinforcementlearning}
Liu, M.; Diao, S.; Lu, X.; Hu, J.; Dong, X.; Choi, Y.; Kautz, J.; and Dong, Y.
  2025.
\newblock ProRL: Prolonged Reinforcement Learning Expands Reasoning Boundaries
  in Large Language Models.
\newblock arXiv:2505.24864.

\bibitem[{Luo et~al.(2026)Luo, Huang, Guo, He, Zou, Hua, and
  Zhang}]{luo2026learningsyntheticdatamodel}
Luo, X.; Huang, Y.; Guo, K.; He, P.; Zou, C.; Hua, T.; and Zhang, X. 2026.
\newblock Learning from Synthetic Data without Model Collapse in Iterative
  Instruction Tuning.
\newblock arXiv:2607.17043.

\bibitem[{Mahrooghi, Lotfi, and
  Abbe(2026)}]{mahrooghi2026goldilocksrltuningtask}
Mahrooghi, I.; Lotfi, A.; and Abbe, E. 2026.
\newblock Goldilocks RL: Tuning Task Difficulty to Escape Sparse Rewards for
  Reasoning.
\newblock arXiv:2602.14868.

\bibitem[{Petrov et~al.(2025)Petrov, Dekoninck, Baltadzhiev, Drencheva,
  Minchev, Balunović, Jovanović, and
  Vechev}]{petrov2025proofbluffevaluatingllms}
Petrov, I.; Dekoninck, J.; Baltadzhiev, L.; Drencheva, M.; Minchev, K.;
  Balunović, M.; Jovanović, N.; and Vechev, M. 2025.
\newblock Proof or Bluff? Evaluating LLMs on 2025 USA Math Olympiad.
\newblock arXiv:2503.21934.

\bibitem[{Röpke et~al.(2026)Röpke, Coward, Lupu, Foster, Rocktäschel, and
  Foerster}]{röpke2026dejaqopenendedevolutiondiverse}
Röpke, W.; Coward, S.; Lupu, A.; Foster, T.; Rocktäschel, T.; and Foerster,
  J. 2026.
\newblock D\'ej\`aQ: Open-Ended Evolution of Diverse, Learnable and Verifiable
  Problems.
\newblock arXiv:2601.01931.

\bibitem[{Shao et~al.(2024)Shao, Wang, Zhu, Xu, Song, Bi, Zhang, Zhang, Li, Wu,
  and Guo}]{shao2024deepseekmathpushinglimitsmathematical}
Shao, Z.; Wang, P.; Zhu, Q.; Xu, R.; Song, J.; Bi, X.; Zhang, H.; Zhang, M.;
  Li, Y.~K.; Wu, Y.; and Guo, D. 2024.
\newblock DeepSeekMath: Pushing the Limits of Mathematical Reasoning in Open
  Language Models.
\newblock arXiv:2402.03300.

\bibitem[{Shi et~al.(2026)Shi, Wu, Song, Zhou, and
  Zhao}]{shi2026efficientreinforcementfinetuningadaptive}
Shi, T.; Wu, Y.; Song, L.; Zhou, T.; and Zhao, J. 2026.
\newblock Efficient Reinforcement Finetuning via Adaptive Curriculum Learning.
\newblock arXiv:2504.05520.

\bibitem[{Stockman, Lawson, and
  Werner(2026)}]{stockman2026earthquakenppbenchmarkearthquakeforecasting}
Stockman, S.; Lawson, D.; and Werner, M. 2026.
\newblock EarthquakeNPP: A Benchmark for Earthquake Forecasting with Neural
  Point Processes.
\newblock arXiv:2410.08226.

\bibitem[{Storchak et~al.(2013)Storchak, Di~Giacomo, Bond{\'a}r, Engdahl,
  Harris, Lee, Villase{\~n}or, and Bormann}]{storchak2013iscgem}
Storchak, D.~A.; Di~Giacomo, D.; Bond{\'a}r, I.; Engdahl, E.~R.; Harris, J.;
  Lee, W. H.~K.; Villase{\~n}or, A.; and Bormann, P. 2013.
\newblock Public Release of the {ISC-GEM} Global Instrumental Earthquake
  Catalogue (1900--2009).
\newblock \emph{Seismological Research Letters}, 84(5): 810--815.

\bibitem[{Sundaram et~al.(2026)Sundaram, Quan, Kwiatkowski, Ahuja, Ollivier,
  and Kempe}]{sundaram2026teachingmodelsteachthemselves}
Sundaram, S.; Quan, J.; Kwiatkowski, A.; Ahuja, K.; Ollivier, Y.; and Kempe, J.
  2026.
\newblock Teaching Models to Teach Themselves: Reasoning at the Edge of
  Learnability.
\newblock arXiv:2601.18778.

\bibitem[{Tsoukalas et~al.(2024)Tsoukalas, Lee, Jennings, Xin, Ding, Jennings,
  Thakur, and
  Chaudhuri}]{tsoukalas2024putnambenchevaluatingneuraltheoremprovers}
Tsoukalas, G.; Lee, J.; Jennings, J.; Xin, J.; Ding, M.; Jennings, M.; Thakur,
  A.; and Chaudhuri, S. 2024.
\newblock PutnamBench: Evaluating Neural Theorem-Provers on the Putnam
  Mathematical Competition.
\newblock arXiv:2407.11214.

\bibitem[{Wang et~al.(2026)Wang, Zheng, Bao, Zhang, Zheng, Chen, Zhang, Feng,
  Khan, Sehgal, Rosin, Paturi, Dube, and
  Bergen}]{wang2026ctopenopenaccessuncontaminated}
Wang, J.; Zheng, Y.; Bao, L.; Zhang, H.; Zheng, Q.; Chen, Y.; Zhang, Y.; Feng,
  M.; Khan, M.; Sehgal, A.~K.; Rosin, C.~D.; Paturi, R.; Dube, U.; and Bergen,
  L. 2026.
\newblock CT Open: An Open-Access, Uncontaminated, Live Platform for the Open
  Challenge of Clinical Trial Outcome Prediction.
\newblock arXiv:2604.16742.

\bibitem[{Yu et~al.(2024)Yu, Jiang, Shi, Yu, Liu, Zhang, Kwok, Li, Weller, and
  Liu}]{yu2024metamathbootstrapmathematicalquestions}
Yu, L.; Jiang, W.; Shi, H.; Yu, J.; Liu, Z.; Zhang, Y.; Kwok, J.~T.; Li, Z.;
  Weller, A.; and Liu, W. 2024.
\newblock MetaMath: Bootstrap Your Own Mathematical Questions for Large
  Language Models.
\newblock arXiv:2309.12284.

\bibitem[{Yuan et~al.(2026)Yuan, Chen, Zheng, Li, Feng, Wang, Xiang, Liu, and
  An}]{yuan2026understandingdiversitycollapserlvr}
Yuan, S.; Chen, J.; Zheng, J.; Li, M.; Feng, L.; Wang, D.; Xiang, T.; Liu, T.;
  and An, B. 2026.
\newblock Understanding Diversity Collapse in RLVR via the Lens of
  Overtraining.
\newblock arXiv:2606.15455.

\bibitem[{Yue et~al.(2025)Yue, Chen, Lu, Zhao, Wang, Yue, Song, and
  Huang}]{yue2025doesreinforcementlearningreally}
Yue, Y.; Chen, Z.; Lu, R.; Zhao, A.; Wang, Z.; Yue, Y.; Song, S.; and Huang, G.
  2025.
\newblock Does Reinforcement Learning Really Incentivize Reasoning Capacity in
  LLMs Beyond the Base Model?
\newblock arXiv:2504.13837.

\bibitem[{Zhang et~al.(2026)Zhang, Li, Ma, Qiu, Tao, Wang, and
  Chu}]{zhang2026d2evodualdifficultyawareselfevolution}
Zhang, R.; Li, R.; Ma, Z.; Qiu, W.; Tao, C.; Wang, Y.; and Chu, X. 2026.
\newblock D$^2$Evo: Dual Difficulty-Aware Self-Evolution for Data-Efficient
  Reinforcement Learning.
\newblock arXiv:2605.17037.

\end{thebibliography}
